\documentclass{article}

\usepackage[preprint,nonatbib]{neurips_2022}

\usepackage{algorithm}
\usepackage{algpseudocode}

\usepackage[utf8]{inputenc} 
\usepackage[T1]{fontenc}    
\usepackage{hyperref}       
\usepackage{url}            
\usepackage{booktabs}       
\usepackage{amsfonts}       
\usepackage{nicefrac}       
\usepackage{microtype}      
\usepackage{xcolor}         
\usepackage{graphicx}
\usepackage{amsmath}
\usepackage{booktabs}
\usepackage{multirow}

\title{nnMamba: 3D Biomedical Image Segmentation, Classification and Landmark Detection with State Space Model}

\author{
  Haifan Gong\textsuperscript{1, 2}, Luoyao Kang\textsuperscript{1,2}, Yitao  Wang\textsuperscript{1,2}, 
  Xiang Wan\textsuperscript{1}, 
  Haofeng Li\textsuperscript{1} \thanks{denotes corresponding authors.}\\
  \textsuperscript{1}Shenzhen Research Institute of Big Data \\
  \textsuperscript{2}SSE, The Chinese University of Hong Kong (Shenzhen) \\
}

\begin{document}

\maketitle

\begin{abstract}
In the field of biomedical image analysis, the quest for architectures capable of effectively capturing long-range dependencies is paramount, especially when dealing with 3D image segmentation, classification, and landmark detection. Traditional Convolutional Neural Networks (CNNs) struggle with locality respective field, and Transformers have a heavy computational load when applied to high-dimensional medical images. 
In this paper, we introduce nnMamba, a novel architecture that integrates the strengths of CNNs and the advanced long-range modeling capabilities of State Space Sequence Models (SSMs). Specifically, we propose the Mamba-In-Convolution with Channel-Spatial Siamese learning (MICCSS) block to model the long-range relationship of the voxels. For the dense prediction and classification tasks, we also design the channel-scaling and channel-sequential learning methods. 
Extensive experiments on 6 datasets demonstrate nnMamba's superiority over state-of-the-art methods in a suite of challenging tasks, including 3D image segmentation, classification, and landmark detection. nnMamba emerges as a robust solution, offering both the local representation ability of CNNs and the efficient global context processing of SSMs, setting a new standard for long-range dependency modeling in medical image analysis. All code and data will be made available after acceptance.
Code is available at \url{https://github.com/lhaof/nnMamba}.
\end{abstract}

\section{Introduction}
Biomedical image analysis plays a pivotal role in healthcare, encompassing key tasks such as segmentation, classification, and landmark detection, which have been extensively studied in the literature~\cite{shen2017deep,he2019non}. Image segmentation, in particular, is crucial for converting unstructured biomedical images into structured and informative representations, thereby propelling scientific discovery and clinical applications forward~\cite{huang2022attentive,xu2023asc}. Semantic segmentation is indispensable across various AI-enabled clinical applications, from diagnostic assistance and therapeutic planning to intra-operative support and monitoring tumor evolution~\cite{isensee2021nnu}. Similarly, image based classification~\cite{kim2022transfer,menze2014multimodal,gong2022less} and landmark detection tasks~\cite{xu2019fetal,li2023sdmt} are vital for a range of downstream biomedical applications. For instance, Alzheimer’s disease (AD), a prevalent neurological disorder, can be better managed through classification models that help in understanding its progression~\cite{kang2023visual}. Automated landmark detection plays a critical role in assisting radiologists by providing key reference points, thereby reducing workload and the potential for diagnostic errors~\cite{xu2019fetal, avisdris2022biometrynet}.

In the field, CNNs~\cite{lecun1995convolutional} have been a dominant force, particularly with FCN-based methods~\cite{long2015fully,li2015visual,ronneberger2015u,chen2018encoder} that are adept at hierarchical feature extraction. Transformers, with their origins in NLP and subsequent adaptation for visual tasks via architectures like ViT~\cite{dosovitskiy2021an} and SwinTransformer~\cite{liu2021swin}, excel in global information aggregation. Their integration into CNN frameworks has led to hybrid models such as TransUNet~\cite{chen2021transunet}, UNETR\cite{hatamizadeh2022unetr}, and nnFormer~\cite{zhou2023nnformer}, which significantly enhance the modeling of long-range dependencies.

However, the computational intensity of Transformers, particularly due to the self-attention mechanism's scaling with input size, poses challenges in high-dimensional biomedical image analysis. State Space Sequence Models (SSMs), and specifically Structured State Space (S4) models~\cite{gu2021efficiently}, have emerged as efficient alternatives, delivering cutting-edge results in long-sequence data analysis~\cite{gu2021combining,gu2021efficiently}. The Mamba model~\cite{gu2023mamba} refines the S4 approach by introducing an input-adaptive mechanism, outperforming Transformers in dense data scenarios. Recent works have begun exploring Mamba's application in medical image segmentation~\cite{ma2024u,xing2024segmamba,yang2024vivim} and visual recognition~\cite{zhu2024vision}. Despite these developments, a comprehensive evaluation of Mamba's effectiveness in 3D medical segmentation, classification, and landmark detection tasks remains an open area of investigation.

\begin{figure}
    \centering
    \includegraphics[width=1\textwidth]{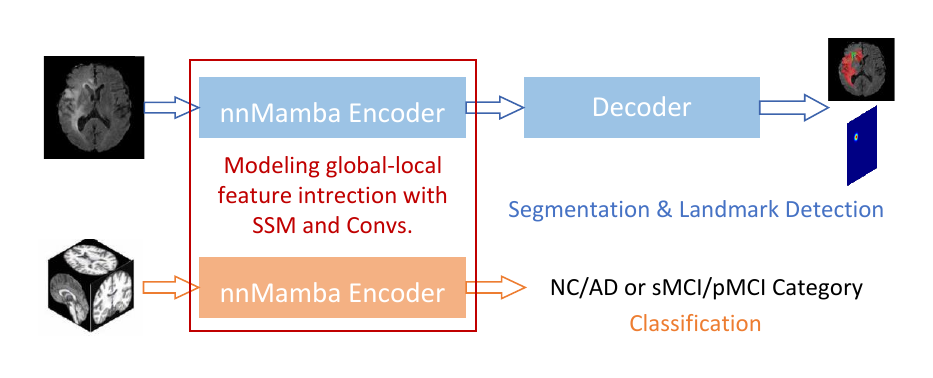}
    \caption{The nnMamba framework is designed for 3D biomedical tasks, focusing on dense prediction and classification. Our approach seeks to tackle the challenge of long-range modeling by leveraging the lightweight and robust long-range modeling capabilities of State Space Models.}
    \label{fig:intro}
\end{figure}

Inspired by the promising results of SSMs in vision tasks and the capabilities of Mamba to handle long sequences efficiently, we explore the integration of Mamba blocks within CNNs to enhance long-range dependency modeling. Thus we proposed the nnMamba, a various of structure for both segmentation, classification, and landmark detection. Inspired by SSMs' efficacy in managing long sequences, our work introduces nnMamba—a versatile architecture designed for these 3D medical imaging applications. The main contributions of our work are summarized below:

\begin{enumerate}
    \item We present nnMamba, a new generation backbone that takes both local and global relationship modeling into account. We established a benchmark comprising six datasets to evaluate backbones across 3 key tasks in 3D medical imaging analysis: segmentation, classification, and landmark detection.

    \item We build the MICCSS (Mamba-In-Convolution with Channel-Spatial Siamese input) module as the basic module in our nnMamba, which enables the long-range relationship modeling ability at both channel and spatial levels.

    \item For dense prediction tasks, we build the encoder with the MICCSS module and use skip scaling to stabilize the training. For the classification task, we add the MICCSS module to the stem layer and tailor-design a hierarchical sequential learning method based on Mamba.
    
    \item Through extensive experimentation, we demonstrate the superior performance of the nnMamba framework over existing methodologies. Our results indicate that nnMamba achieves state-of-the-art effectiveness across the board, setting a new method for future research and applications in the field of 3D medical image analysis.
\end{enumerate}


\begin{figure}[t]
    \centering
    \includegraphics[width=1.02\textwidth]{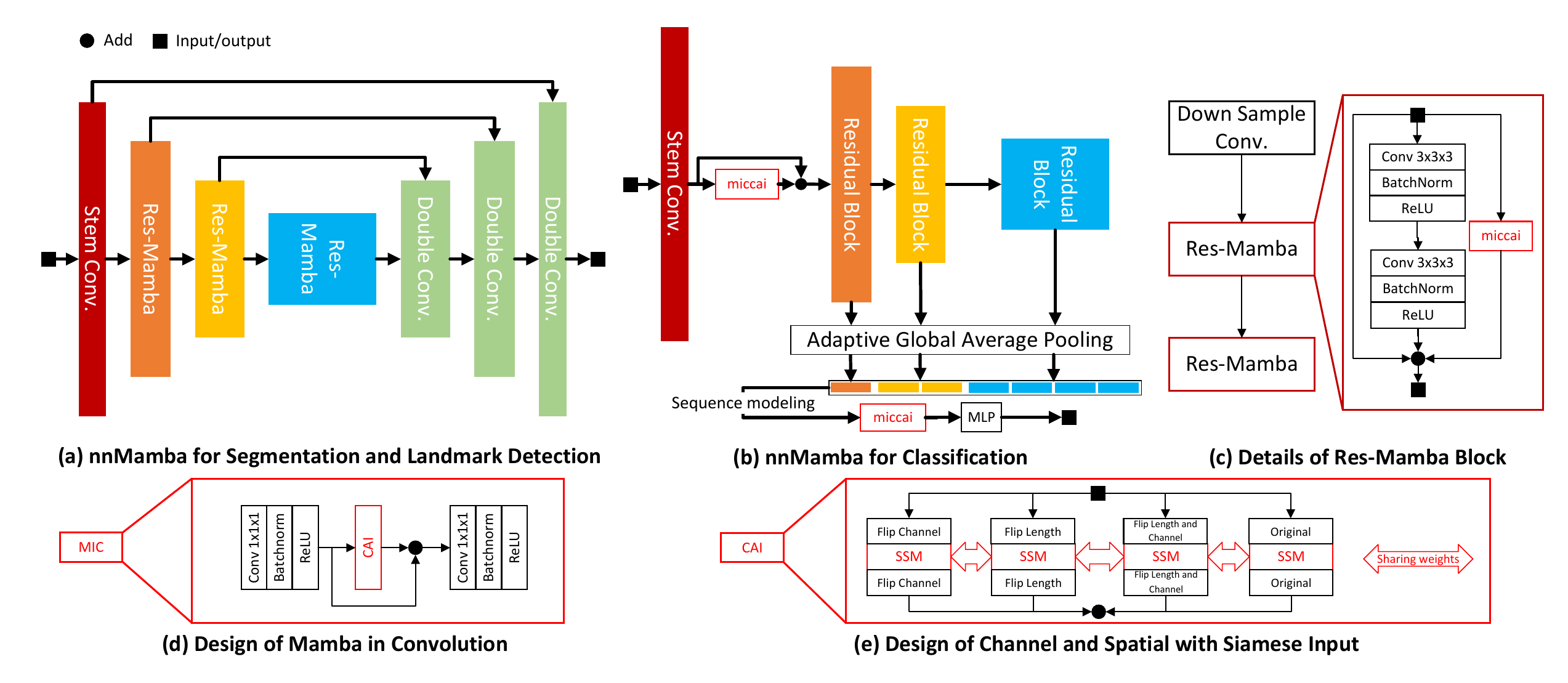}
    \caption{Illustrative diagrams of the nnMamba framework architectures. (a) Presents the network structure for segmentation and landmark detection tasks. (b) Depicts the architecture tailored for classification tasks. Detailed structures of the blocks utilized within our networks are shown in (c), (d), and (e).}
    \label{fig:enter-label}
\end{figure}

\section{Methodology}
We first briefly review the State Space Sequence Models (SSMs) before introducing our nnMamba. SSMs are introduced as mechanisms for mapping one-dimensional input sequences \( u(t) \) to outputs \( y(t) \in \mathbb{R} \) through linear ordinary differential equations:
\begin{equation}
    x'(t) = Ax(t) + Bu(t);\quad y(t) = Cx(t),
\end{equation}
where \( x(t) \in \mathbb{R}^N \), and \( A \in \mathbb{R}^{N \times N} \), \( B, C \in \mathbb{R}^N \) are the system parameters. SSMs suffer from memory and gradient issues, which Structured State Space models address by structuring \( A \) with a projection operator for improved long-range dependency capture. Mamba extends SSMs with adaptive parameter tuning and efficient computation, thus providing enhanced performance on long sequences.

\subsection{MICCSS: Mamba-In-Convolution with Channel-Spatial Siamese Learning}
\textbf{Parallel CNN and SSM Design.} Despite the success of sequential modeling in natural image processing~\cite{liu2021swin}, its application in medical imaging is challenged by the limited sample sizes (typically less than 1,000)~\cite{isensee2021nnu} compared to those available for natural images (over 1 million in ImageNet). This makes it difficult to achieve performance comparable to Convolutional Neural Networks (CNNs). Additionally, CNNs' inductive biases are suited to medical imaging, where patterns often consist of similar intensity repeats in local regions. Thus, we introduce Mamba as a parallel module enhancing CNNs' ability to model long-range dependencies, which is shown in Fig.2~(c).

\noindent \textbf{MIC: Rethinking Network-In-Network.} Drawing inspiration from Network-In-Network (NIN)~\cite{Lin2013nin}, we propose to first filter out the useful features through the conventional layers, then send the feature to mamba for better non-linear long-range modeling. In detail, we develop the Mamba-In-Convolution (MIC) module, formalized as:
\begin{equation}
    \mathbf{F}_{\text{out}} = \text{Convs.O}(\text{SSM}(\text{Convs.I}(\mathbf{F}_{\text{in}})) + \text{Convs.I}(\mathbf{F}_{\text{in}})).
\end{equation}
As depicted in Fig.~2(d), the MIC block integrates SSM with Convolutional layers, denoted as \( \text{Convs.O} \) and \( \text{Convs.I} \), which include a \( 1 \times 1 \) convolution, batch normalization, and ReLU activation. Feature maps of shape \( B \times C \times H \times W \times D \) are initially processed to filter features, then reshaped to \( B \times C \times L \) for global interaction modeling via the SSM.

\noindent \textbf{CSS: Leveraging SSM for Channel and Spatial Feature Interaction.} The core strength of the Network-In-Network (NIN) approach lies in the intermediate MLP layer's nonlinear capacity to encapsulate features. However, this MLP layer struggles with flattened features of dimensionality \( C \times H \times W \times D \). To address this, we investigate the potential of State Space Models (SSMs), which are inherently adept at modeling long-range relationships within sequences. Drawing inspiration from Siamese Networks\cite{bertinetto2016fully,gong2023unbiased}, we conceive the Channel-Spatial Siamese (CSS) input module, which fully harnesses the representational capabilities of SSMs. In this design, we augment the reshaped input feature of dimension \( B \times C \times L \) across both channel and spatial dimensions. These augmented features are then processed by an SSM with shared weights, as detailed in Algorithm 1.

\begin{algorithm}
\caption{CSS: Channel-Spatial Siamese inputting for long-range modeling}
\begin{algorithmic}[1]
\State SiamSSM \Comment{SSM with shared parameters} 
\State $x_{\text{flat}} \gets \text{input feature with shape } [B, L, C]$
\State $x_{\text{mamba}} \gets \text{SiamSSM}(x_{\text{flat}})$
\Comment{Iterate over combinations of flipping dimensions}
\For{$d \text{ in } \{\text{[1]}, \text{[2]}, \text{[1, 2]}\}$}
    \State $x_{\text{flip}} \gets \text{flip}(x_{\text{flat}}, \text{dims}=d)$
    \State $x_{\text{mamba}} \gets x_{\text{mamba}} + \text{flip}(\text{SiamSSM}(x_{\text{flip}}), \text{dims}=d)$
\EndFor
\State $x_{\text{mamba}} \gets \frac{1}{4} x_{\text{mamba}}$
\end{algorithmic}
\end{algorithm}

\subsection{nnMamba for Segmentation and Landmark Detection}
For dense prediction tasks such as segmentation or heatmap-based landmark detection, we implement a UNet architecture to capture the long-range dependencies between the encoder and decoder. Our framework utilizes a residual-based encoder paired with a convolutional decoder. We maintain the same architectural structure for both segmentation and landmark detection tasks. Notably, to stabilize the training of the UNet, we apply a learning-based scaling method~\cite{huang2022scale} during the concatenation operation. Let \( X_h \) be the feature map from the \( h \)-th block of the encoder, \( X_h' \) be the corresponding feature map from the decoder, and \( X_{\text{cat}} \) be the post-concatenation feature map. Let \( SE \) denote the squeeze-extraction module as described in~\cite{huang2022scale}. The process is formulated as:
\begin{equation}
    X_{\text{cat}} = [X_h', X_h \cdot SE(\text{pooling}(X_h))].
\end{equation}

\subsection{nnMamba for Classification}
For classification tasks, we adopt a residual encoder-centric architecture. We posit that reducing the size of feature maps amplifies the ability of convolutional operations to capture long-range dependencies. With this in mind, we strategically incorporate the Mamba layer early in the network, immediately following initial feature extraction. Positioning the Mamba layer at this stage ensures that features are endowed with global context early on, potentially diminishing the necessity for complex operations later in the network for abstract reasoning.

To further refine supervision across various feature map resolutions, we propose treating hierarchical features as a sequence, which are then processed by the Mamba to extract multi-level characteristics. Specifically, let \( X_h \) (for \( h = 2, 3, 4 \)) denote the feature map from the second to the fourth block. We first apply a max-pooling operation to these feature maps, yielding \( P_h \) with a shape of \( B \times C \times 2^h \). These are then reshaped into a sequence with dimensions \( B \times 2C \times (1 + 2 + 3) \). Lastly, the SSM is employed to distill the hierarchical sequential features, which are subsequently fed into the MLP.

\section{Experiments}
\subsection{Implementation and Metric}
All the models are trained with NVIDIA V100 GPU with 24GB memory. The framework is implemented in PyTorch 2.0.1 and CUDA 11.6. We train the models with the Adam optimizer at a learning rate of 0.002, batch size at 2, training epoch at 1000, and weight decay at 0.001 for segmentation tasks. For other tasks, we only train 100 epochs. For the segmentation task, we use the Dice, NSD, and HD95 for evaluation by following\cite{reinke2024understanding}. For the landmark detection task, we take the MRE as the metric for evaluation following \cite{li2023sdmt}. For the classification task, we use the AUC, Accuracy, and F1-score by following \cite{kang2023visual}. Details of the datasets and data split used in this work are available in the supplementary material. If there is a validation set, we select the best-performed model on the validation set and evaluate it on the test set. Otherwise, the results are obtained by averaging the results of three different seeds. The visualization results are also available in the appendix.

\subsection{Segmentation Evaluation}
Table~\ref{tab:brats2023gil} illustrates the competitive performance of nnMamba in comparison to other state-of-the-art methods on the BraTS 2023 GIL task \cite{menze2014multimodal}. nnMamba consistently achieves top-tier results across all evaluated metrics for brain tumor segmentation, demonstrating its robustness and accuracy. The method's effectiveness is further highlighted by its performance in the challenging Hausdorff distance metric, showcasing its precision in delineating tumor boundaries. It can be seen that our method can significantly boost the performance on the boundary modeling of the tumor, especially for the long-range relationship modeling of enhanced tumors.

In Table~\ref{tab:andi}, nnMamba showcases superior performance on the AMOS2022 dataset\cite{ji2022amos}, particularly in the MRI-Test category, where it achieves a mean Dice coefficient (mDice) of 73.98\% and a mean Normalized Surface Dice (mNSD) of 65.13\%, outperforming all other compared methods. It might be our lightweight design and long-range modeling to boost the performance. Notably, nnMamba achieves these results with significantly lower parameters and computational complexity, boasting only 15.55 MB parameters and 141.14 GFLOPS. These results underline nnMamba's efficiency and effectiveness in medical image segmentation tasks.

\begin{table}[]
\centering
\setlength{\tabcolsep}{5pt}
\caption{Evaluation of nnMamba and other state-of-the-art methods on the BraTS 2023 GIL track.}
\begin{tabular}{@{}c|cccc|cccc@{}}
\toprule
BraTS 2023 GIL & \multicolumn{4}{c|}{Dice}       & \multicolumn{4}{c}{HD95}      \\ \midrule
Methods        & WT    & TC    & ET    & Average & WT   & TC   & ET    & Average \\ \midrule
UniMiss\cite{xie2022unimiss}& 93.48& 90.06& 84.40& 89.31& 4.55& 6.80& 13.70& 8.38\\
DIT\cite{peebles2023dit}            & 93.49 & 90.22 & 84.38 & 89.36   & 4.21 & 5.27 & 13.64 & 7.71    \\
UNETR\cite{hatamizadeh2022unetr}          & 93.33 & 89.89 & 85.19 & 89.47   & 4.76 & 7.27 & 12.78 & 8.27    \\
nnUNet\cite{isensee2021nnu}         & 93.31 & 90.24 & 85.18 & 89.58   & 4.49 & 4.95 & 11.91 & 7.12    \\
nnMamba        & 93.46 & 90.74 & 85.72 & 89.97   & 4.18 & 5.12 & 10.31 & 6.53\\ \bottomrule
\end{tabular}
\label{tab:brats2023gil}
\end{table}

\begin{table}[]
\centering
\setlength{\tabcolsep}{5pt}
\caption{Evaluation of nnMamba on the AMOS2022 dataset~\cite{ji2022amos}. The results of the test sets are obtained by submitting them to the official online testing site.}
\begin{tabular}{@{}ccccccc@{}}
\toprule
AMOS       & \multicolumn{2}{c}{Effiency}  & \multicolumn{2}{c}{CT-Test}  & \multicolumn{2}{c}{MRI-Test} \\ 
Metrics    & Parameter(M)& Flops(G)& mDice         & mNSD         & mDice         & mNSD         \\ \midrule
nnUNet\cite{isensee2021nnu}& 31.18          & 680.31       & 89.04         & 78.32        & 67.63         & 59.02        \\
VNet\cite{milletari2016vnet}      & 45.65          & 849.96       & 82.92         & 67.56        & 65.64         & 57.37        \\
CoTr \cite{xie2021cotr}      & 41.87          & 668.15       & 80.86         & 66.31        & 60.49         & 51.18        \\
nnFormer\cite{zhou2023nnformer}   & 150.14         & 425.78       & 85.61         & 72.48        & 62.92         & 54.06        \\
UNETR\cite{hatamizadeh2022unetr}      & 93.02          & 177.51       & 79.43         & 60.84        & 57.91         & 47.25        \\
SwinUNetr\cite{hatamizadeh2021swin} & 62.83          & 668.15       & 86.32         & 73.83        & 57.50         & 47.04        \\
nnMamba    &15.55             &  141.14  & 89.63   & 79.73       &73.98  &  65.13\\ \bottomrule
\end{tabular}
\label{tab:andi}
\end{table}

\subsection{Classification Evaluation}
We evaluate the classification task on the ADNI dataset\cite{jack2008alzheimer,lian2018hierarchical} with the AD and NC classification task. We use the transfer learning setting in \cite{kang2023visual} to evaluate the methods by first training the networks on the NC AD classification task, then we transfer the weights and make the evaluation on the sMCI pMCI classification task.
Table~\ref{tab:andi} shows the performance of nnMamba and benchmark methods on the ANDI classification task, comparing non-converters (NC) to Alzheimer's Disease (AD) and stable Mild Cognitive Impairment (sMCI) to progressive MCI (pMCI). In both tasks, nnMamba outperforms the other models, achieving the highest Accuracy (ACC), F1 score, and Area Under the Curve (AUC) metrics. The results indicate nnMamba's effectiveness, particularly in the NC VS AD task with an ACC of \(89.41\%\pm0.85\), an F1 score of \(88.68\%\pm0.77\), and an AUC of \(95.81\%\pm0.59\). Moreover, these results also show that our method can successfully boost the performance of transferring the learned weights to the downstream tasks on classification.

\begin{table}[]
\centering
\caption{Evaluation of nnMamba and other state-of-the-art methods on the ANDI classification task.}
\setlength{\tabcolsep}{3pt}
\begin{tabular}{@{}c|ccc|ccc@{}}
\toprule
Task     & \multicolumn{3}{c|}{NC VS AD}                             & \multicolumn{3}{c}{sMCI VS pMCI}                          \\ \midrule
Methods  & ACC               & F1                & AUC               & ACC               & F1                & AUC               \\ \midrule
ResNet \cite{he2016deep}  & 88.40$_{\pm3.41}$  & 88.00$_{\pm2.81}$ & 94.93$_{\pm0.72}$ & 67.96$_{\pm1.50}$ & 52.14$_{\pm1.51}$ & 74.94$_{\pm2.18}$ \\
DenseNet \cite{huang2017densely} & 87.95$_{\pm0.70}$ & 86.93$_{\pm0.87}$ & 94.86$_{\pm0.40}$ & 73.12$_{\pm3.10}$ & 53.30$_{\pm2.99}$ & 76.31$_{\pm3.09}$ \\
ViT \cite{dosovitskiy2021an}    & 88.85$_{\pm1.17}$ & 87.66$_{\pm1.72}$ & 94.12$_{\pm1.29}$ & 67.16$_{\pm3.16}$ & 51.68$_{\pm5.72}$ & 75.08$_{\pm6.88}$ \\
CRATE \cite{yu2023white} & 84.69$_{\pm2.53}$ & 82.66$_{\pm3.47}$ & 91.42$_{\pm1.43}$ & 70.63$_{\pm2.60}$ & 53.41$_{\pm2.53}$ & 76.06$_{\pm2.98}$ \\
nnMamba  &  89.41$_{\pm0.85}$  &  88.68$_{\pm0.77}$  &  95.81$_{\pm0.59}$ & 75.79$_{\pm1.79}$ & 56.55$_{\pm2.37}$ &   76.84$_{\pm0.84}$ \\ \bottomrule
\end{tabular}
\end{table}

\subsection{Landmark Detection Evaluation}
For the landmark detection task, we evaluate our method on our private fetal cerebellum landmark detection dataset. Table~\ref{tab:landmark} compares the performance of nnMamba with ResUNet and VitPose on a landmark detection task across six test conditions (TCD 1, TCD 2, HDV 1, HDV 2, ADV 1, ADV 2) and reports the average error across all conditions. nnMamba consistently exhibits lower error rates than the other methods in individual test conditions and maintains the lowest average error at 2.11. This underscores nnMamba's superior accuracy in landmark detection compared to the state-of-the-art methods listed.


\begin{table}[]
\centering
\caption{Evaluation of nnMamba and other state-of-the-art methods on the landmark detection task.}
\begin{tabular}{@{}cccccccc@{}}
\toprule
Landmark Detection & TCD 1 & TCD 2 & HDV 1 & HDV 2 & ADV 1 & ADV 2 & Average \\ \midrule
ResUNet\cite{he2016deep}            & 2.23  & 2.15  & 2.32  & 2.16  & 2.45  & 2.05  & 2.23    \\
VitPose\cite{xu2022vitpose}	        & 3.25 	& 3.24 	& 2.98 	& 3.34  & 3.32  & 2.59  & 3.12      \\
nnMamba            & 2.21  & 2.18  & 2.10  & 2.14  & 2.06  & 1.99  & 2.11 \\ \bottomrule
\end{tabular}
\label{tab:landmark}
\end{table}

\subsection{Ablation Study}
Table~\ref{tab:ablationstudy} presents the results of an ablation study, demonstrating the incremental benefits of MIC and MICCSS on various medical image analysis tasks. The integration of MIC and MICCSS into the baseline method yields consistent enhancements across segmentation metrics on the Brats2023 dataset, as evidenced by improved Dice and HD95 scores. Additionally, in landmark detection, the incorporation of these techniques leads to a decreased MRE, signifying more accurate localization capabilities. The ADNI classification task also shows a trend of increased AUC scores with each added component, highlighting the potential of MIC and MICCSS to boost the performance of classification models.
\begin{table}[]
\centering
\setlength{\tabcolsep}{6pt}
\caption{ The ablation study on through evaluation metrics of segmentation, landmark detection, and classification tasks. ``MIC'' indicates directly using the SSM module.}
\begin{tabular}{c|cc|c|cc}
\toprule
Methods& \multicolumn{2}{c|}{Brats2023} &  Landmark& \multicolumn{2}{c}{ADNI Classification}\\
 Metric& Dice& HD95& MRE& -&AUC\\ \midrule
Baseline  &89.58&7.12& 2.23 & Baseline & 94.93   \\
+MIC      &89.92&6.94& 2.14 & +MICCSS & 95.35   \\
+MICCSS   &89.97  &6.53     & 2.11 & +Sequence & 95.81  \\
 \bottomrule
\end{tabular}
\label{tab:ablationstudy}
\end{table}

\subsection{Visualization Results}
We also provides the visualization results in Figure 3 and Figure 4, from these figures, we can observe that our method can get better results by modeling long range relationship.

\begin{table}[]
\centering
\caption{Details of the dataset used in our study.}
\begin{tabular}{@{}cccclccc@{}}
\toprule
Task                            & Dataset & Dataset Type & Modality & Spacing            & Train & Valid. & Test \\ \midrule
\multirow{3}{*}{Segmentation}   & Brats   & Tumor        & MR       & $1\cdot 1\cdot 1 mm^3$       & 875      & 125        & 251     \\
                                & AMOS-CT & Organ        & CT       & $1.25\cdot 5\cdot 2 mm^3$    & 200      & 100        & 200     \\
                                & AMOS-MR & Organ        & MR       & $0.86\cdot 3\cdot 3 mm^3$    & 40       & 20         & 40      \\ \midrule
\multirow{2}{*}{Classification} & ADNI    & AD/NC        & MR       & $1\cdot 1\cdot 1 mm^3$       & 351      & -          & 296     \\
                                & ADNI    & sMCI/pMCI    & MR       & $1\cdot 1\cdot 1 mm^3$       & 356      & -          & 337     \\ \midrule
Landmark                        & Private & Landmark     & MR       & $0.6\cdot 0.6\cdot 0.6 mm^3$ & 120      & -          & 60      \\ \bottomrule
\end{tabular}
\end{table}

\begin{figure}
\centering
\includegraphics[width=1\textwidth]{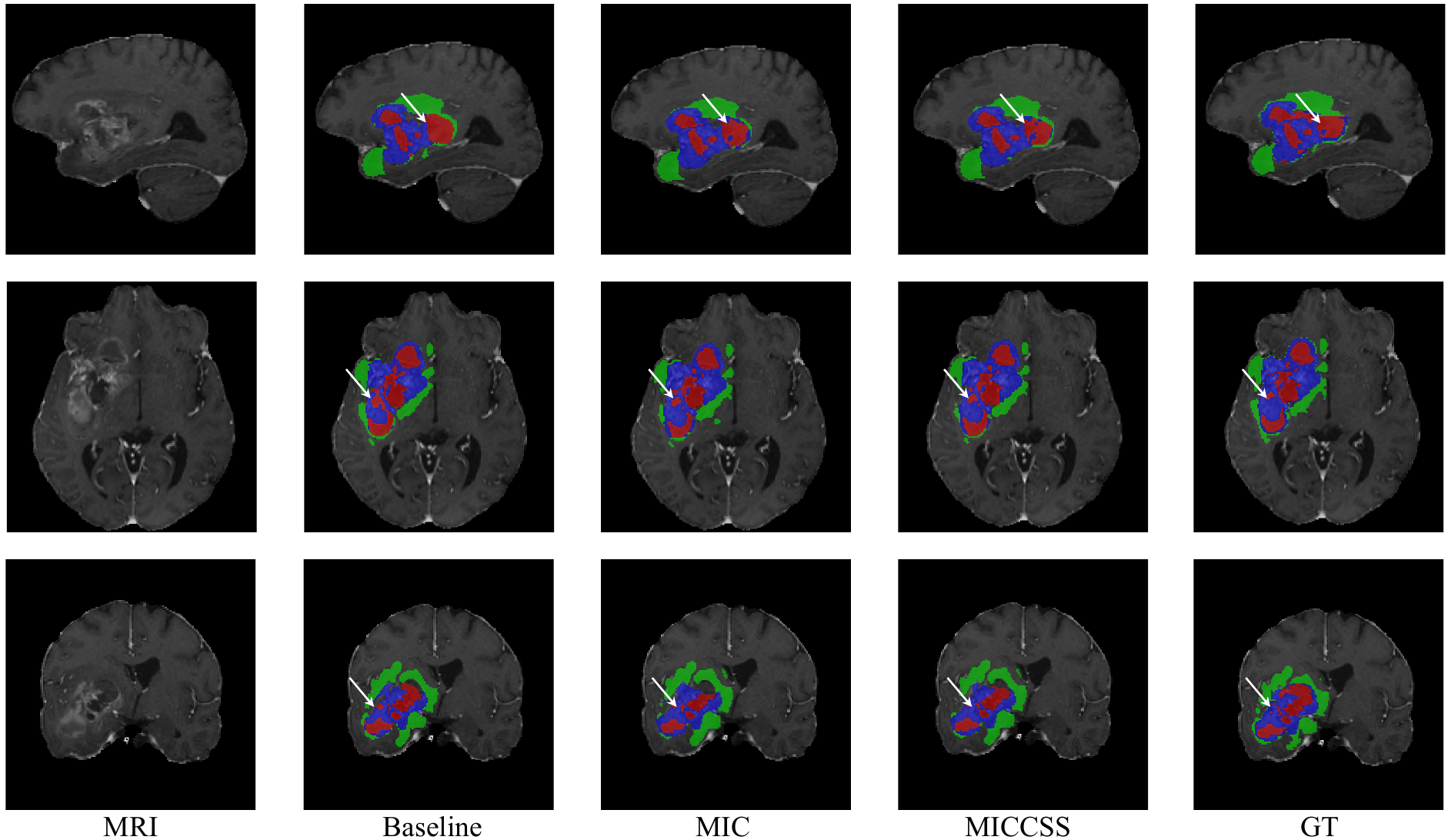}
\caption{Visualization of segmentation predictions on the BraTS2023 dataset. From left to right, the columns display the MRI image, the prediction result of nnUNet, the prediction result of our nnMamba (MIC) model, the prediction result of our nnMamba (MICCSS) model, and the ground truth segmentation. This side-by-side comparison shows that our model can better capture the discontinue region of the tumor.}
\label{vis}
\end{figure}

\begin{figure}
\centering
\includegraphics[width=1\textwidth]{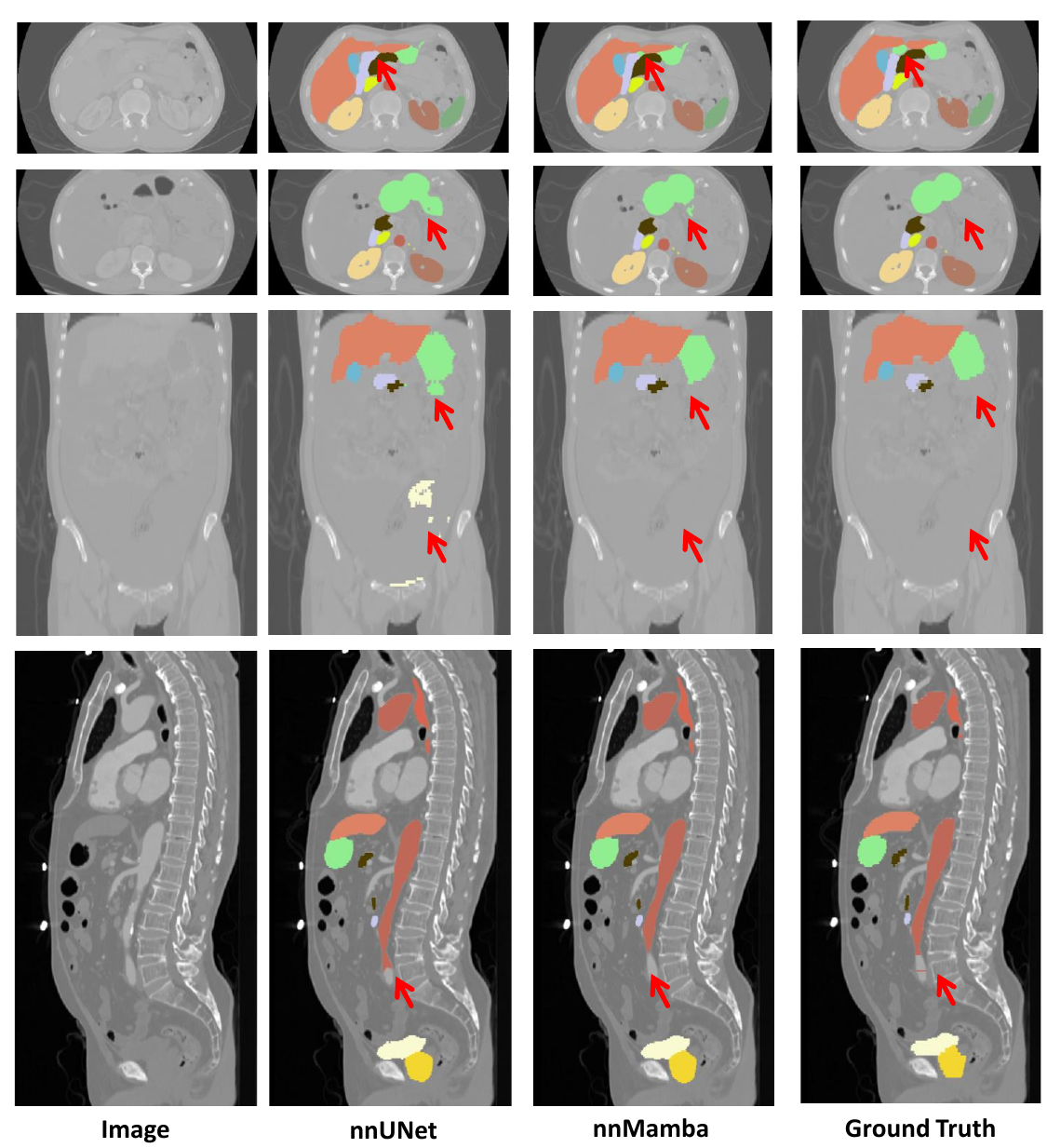}
\caption{Visualization and comparison of segmentation predictions on the AMOS22 CT validation dataset. The four columns display, respectively, the CT image, the ground truth segmentation, the prediction result of nnUNet, and the prediction result of our nnMamba model. By capturing long-range dependencies, our nnMamba model effectively reduces over-segmentation and missed segmentation issues, particularly over long distances.}
\label{vis}
\end{figure}

\section{Conclusion}
Our work presents nnMamba, a new framework combining the detailed feature extraction of CNNs with the broad dependency modeling of SSMs. Designed for 3D medical image analysis, nnMamba excels in segmentation, classification, and landmark detection. It outperforms current methodologies, offering a blend of local and global image context understanding. nnMamba's robust performance has significant implications for improving medical diagnostics and therapies.

\section*{Acknowledgments and Disclosure of Funding}
This work was supported in part by the National Natural Science Foundation of China (NO.~62102267), and in part by the Guangdong Basic and Applied Basic Research Foundation (2023A1515011464). 

\bibliographystyle{unsrt} 
\bibliography{main}


\appendix



\end{document}